\documentclass[11pt]{article}

\usepackage{arxiv}
\usepackage[utf8]{inputenc} 
\usepackage[T1]{fontenc}    
\usepackage{hyperref}       
\usepackage{url}            
\usepackage{booktabs}       
\usepackage{amsfonts}       
\usepackage{nicefrac}       
\usepackage{microtype}      
\usepackage{lipsum}
\usepackage{orcidlink}
\usepackage{amsmath}
\usepackage{graphicx}
\graphicspath{ {./images/} }

\title{GeoAI-based post-segmentation quality validation of building footprints via spatial feature engineering}

\author{
 Shah Imran Ahsan Chowdhury\orcidlink{0009-0003-6928-3889} \\
  Department of Computer Science and Engineering \\
  Jahangirnagar University\\
  Dhaka 1342, Bangladesh \\
  \texttt{imranahsanchowdhury@gmail.com} \\
    \And
  Kazi Jihadur Rashid\orcidlink{0000-0001-9524-825X} \\
  Department of Geography\\
  Florida State University\\
  Tallahassee, FL 32306, USA \\
  \texttt{kr24x@fsu.edu} \\
 \And
  Rajsree Das Tuli\orcidlink{0000-0002-2916-2324}\\
  Department of Geography and Environmental Sustainability\\
  The University of Oklahoma\\
  Norman, OK 73019, USA \\
  \texttt{rajsree.tuli@ou.edu} \\
  \And
 Rahul Saha \\
  Urban Development Directorate\\
  Dhaka, Bangladesh \\
  \texttt{rahulsaha2626@gmail.com} \\
  \And
 Bulbul Ahammad \\
  Department of Computer Science and Engineering \\
  Jahangirnagar University\\
  Dhaka 1342, Bangladesh \\
  \texttt{bulbul@juniv.edu} \\
}

\begin{document}
\maketitle
\begin{abstract}
Deep learning-based building footprint extraction from high-resolution imagery often produces topologically inconsistent vectors unfit for direct GIS database ingestion. To address this, we present a multidomain GeoAI quality control framework that automates error detection to systematically purify vector footprint databases. Candidate footprints were generated across five UAV survey sites in Bangladesh using U-Net (ResNet-34) and SAM-LoRA (ViT-B). The extracted raster masks were vectorized, geometrically regularized, and consolidated under a spatial-exclusivity constraint to eliminate duplicate representations. We used twenty-four predictors capturing geometric, spatial-contextual, and raster-derived spectral and texture properties. Machine Learning (ML) classifiers were trained on a development partition (Sites B–D) and rigorously validated on a spatially independent test set (Site E) excluded from hyperparameter tuning and class balancing. The experimental results demonstrate that geometric and spatial-contextual predictors using Decision Tree (DT) provide the most effective discriminatory evidence for identifying object-level boundary deformations. DT achieved an accuracy of 95.31\%, an F1-score of 91.06\%, and a Matthews correlation coefficient (MCC) of 0.880 on the unseen testing site. At the database level, this framework successfully identified 87.34\% of erroneous footprints while maintaining 98.31\% of acceptable structures, reducing the residual error proportion from 27.32\% to 4.62\% and improving final database purity to 95.38\%. This translates into a relative error reduction of 83.09\%. The findings indicate that post-segmentation object-level ML provides a highly transferable, robust mechanism for automated quality assurance in production-ready geographic information system (GIS) workflows.
\end{abstract}


\section{Introduction}
As geospatial data have become more widely available, the need for accurate and up-to-date information about the built environment has grown substantially. This demand for detailed built environment has pushed automated building footprint extraction to the forefront of remote sensing research. Accurate and precise infrastructure data is essential for a wide array of Geospatial workflows ranging from applications such as topographic mapping, land administration, urban planning, disaster management, and 3-D city modeling (Awrangjeb et al., 2020; Blaschke, 2010; Gilani et al., 2016; Luo et al., 2021). Despite steady advances in computer vision and deep learning, consistent footprint extraction remains highly challenging. One of the challenges arises from the large diversity of buildings. Buildings differ considerably in size, shape, roof structure, construction materials, and spatial arrangement across different geographic regions. The problem becomes even more complex when buildings are partially hidden by shadows or vegetation, located in densely built-up areas, or have spectral characteristics similar to nearby impervious surfaces such as roads and parking lots (Fenglei et al., 2025; Hong et al., 2023; Liu et al., 2019). These challenges are somewhat more evident in very high-resolution imagery. The finer the details, the more complex the building delineation.\\
Deep learning models have significantly improved building detection and semantic segmentation performance in current times. However, accurate pixel-level classification does not necessarily translate into geometrically precise building footprints. Many existing methods still struggle to accurately identify building boundaries, preserve building geometry, and separate neighboring structures. As a result, the extracted footprints often contain fragmented, merged, or distorted polygons (Hong et al., 2023; Marcos et al., 2018; Zhao et al., 2018). Such geometric inaccuracies limit the practical use of the generated footprints in GIS databases, cadastral mapping, engineering applications, and other spatial analysis workflows. Moreover, the model's efficiency can be significantly decreased while applying a trained model to different sensors, geographic locations, or types of urban form because of differences in image characteristics and training label quality (Luo et al., 2021; Vats et al., 2024). Therefore, future research should place greater emphasis on generating building footprints that are not only accurate but also geometrically precise.\\
Building Footprint Extraction (BFE) methods have evolved considerably in the past decade. From conventional feature engineering techniques, it has moved to more sophisticated deep learning architectures, which can process increasing complexity of urban landscapes. Early studies largely relied on object-based image analysis and manually designed features to characterize building characteristics in remote sensing imagery. For instance, according to Chen et al. (2018), object-based features such as the Edge Regularity Indices (ERI) and Shadow Line Indices (SLI) improved building detection in high-resolution RGB imagery by exploiting geometric regularity and shadow information. Several review studies have highlighted that combining geometric information with semantic features often leads to more accurate building extraction results.\\
Abdollahi et al. (2020) introduced a GAN-based framework designed to generate building footprints with more realistic geometric characteristics. Dabove et al. (2024) demonstrated that integrating digital surface models (DSMs) with RGB imagery improves building prediction accuracy by incorporating height information. Hu et al. (2021) developed the Deep Automatic Building Extraction Network (DABE-Net), combining attention mechanisms and recurrent layers to refine segmentation boundaries. However, these model architectures are quite complicated and often require substantial computational resources. Tejeswari et al. (2022) argued that limited availability of location-specific training data remains a greater challenge than model architecture for producing reliable building footprints. To address computational limitations, He et al. (2024) proposed a lightweight semantic segmentation framework capable of recovering building boundaries in complex urban environments while maintaining computational efficiency. Recent advancements in building extraction research have been focused more and more on preserving the integrity of the buildings under difficult scenarios like vegetation occlusions and shadow interference. For example, Tang et al. (2024) developed BIENet, a framework that leverages structural reasoning to reconstruct complete building geometry from fragmented or partial observations.\\
Despite these advancements, critical bottlenecks still persist. Even now, most deep learning frameworks still treat footprint extraction as a standard pixel-wise semantic segmentation task. These models generate raw raster masks that require extensive post-processing such as vectorization and geometric regularization in order to integrate into a GIS database (Hu et al., 2021; Li et al., 2024). While recently introduced polygon-aware neural networks, contour vectorization algorithms, and topology-preserving regularization methods allow us to better represent boundaries of building footprints (Bulatov et al., 2024; Li et al., 2023; Sanca et al., 2023; Schuegraf et al., 2024; Zhang et al., 2024), better geometric structure does not always guarantee correct object representations. The regularized footprint may still contain false positives, false negatives, merged neighbors, object fragments, or local deformations. In addition, current building extraction models have limited capabilities of generalization across different types of urban landscapes and imaging conditions due to different sensor and imaging parameters and annotation data quality(Li et al., 2024; Vats et al., 2024; Yuan et al., 2025). Thus, despite a lot of progress in increasing the accuracy of segmentations, relatively few studies were dedicated to a systematic evaluation of the extracted footprints' reliability.\\
However, recent research has shown that assessing the quality of the geospatial data generated by GeoAI methods is not enough when using traditional measures based on overlaps since multiple aspects of geometrical, positional, and semantic correctness should be considered (Niroshan and Carswell, 2025). Also, object-based remote sensing methods have proven that geometric, context, texture, and image properties offer complimentary data for classification (Shen et al., 2023). Furthermore, the importance of geographically independent validation was proven as a more realistic measure of model transferability than random partitioning because of overestimation caused by spatial autocorrelation (Koldasbayeva et al., 2024; Wang et al., 2023). Nevertheless, a unified framework capable of learning recurrent footprint errors from multiple extraction architectures, eliminating duplicated predictions prior to model development, evaluating quality classifiers under spatially independent conditions, and translating classification outcomes into measurable improvements in building footprint databases has received little attention.\\
To overcome these limitations, this study develops a post-segmentation GeoAI quality assurance filter bridging deep learning outputs with GIS vector database standards. Rather than comparing segmentation architectures, a CNN-based U-Net and a transformer-based SAM-LoRA are used as complementary sources of heterogeneous footprint candidates, enabling the quality assessment framework to learn recurrent errors generated by fundamentally different extraction paradigms. Segmentation outputs are converted into vector footprints, geometrically regularized, and characterized using twenty-four geometric, spatial-contextual, and raster-derived features. These object-level descriptors are subsequently used to train and evaluate ML classifiers under multiple feature-domain combinations. Moreover, rigorous validation on a spatially disjoint study site ensures transferability across varying urban morphologies. Overall, this semi-automated GeoAI quality assessment framework improves the quality and reliability of building footprint databases.

\section{Materials and Methods}
\label{sec:headings}
\begin{figure}[htbp]
  \centering
  \includegraphics[width=0.85\linewidth]{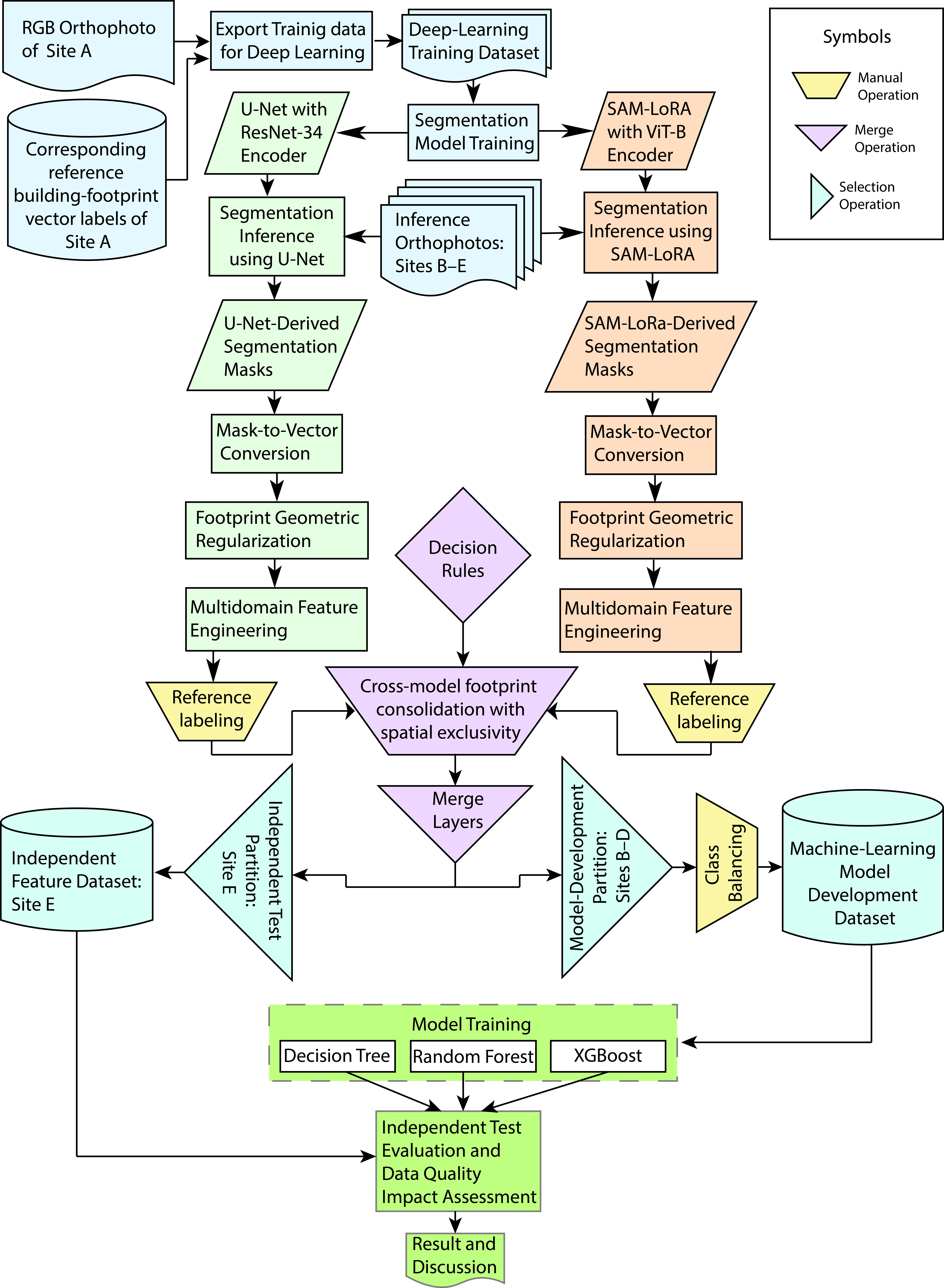}
  \caption{GeoAI workflow for UAV building-footprint extraction and post-segmentation quality classification, showing training-data export, dual-model inference, mask-to-vector processing, feature engineering, cross-model footprint consolidation under a spatial-exclusivity constraint, partitioning, class balancing, model training, and independent testing.}
  \label{fig:fig1}
\end{figure}

The workflow combined deep-learning (DL) semantic segmentation with object-level machine-learning (ML) analysis of UAV-derived building footprints ((\autoref{fig:fig1})). We exported the RGB orthophoto from Site A along with its reference building-footprint vector labels to create the DL training dataset. It was done for both U-Net with ResNet-34 and SAM-LoRA with ViT-B backbone models. Then, we applied these trained models independently to orthophotos from Sites B-E for inference. The resulting predicted masks were transformed into vector footprints, which were geometrically regularized and enhanced with 24 predictors derived from geometric, spatial-contextual, and raster data in distinct model-specific layers. A cross-model footprint consolidation was executed under a spatial-exclusivity constraint prior to merging layers at the site level. For U-Net and SAM-LoRA polygons representing the same visible building, a single representative footprint was chosen based on orthophoto-based decision rules outlined in Section 3.5; if only one model detected a building, its polygon was retained. The merged feature layers were divided into a model-development set from Sites B-D and a spatially independent test set from Site E, with class balancing applied solely to the model-development partition before training ML models.  

\subsection{Study area and data sources}
\begin{figure}[htbp]
  \centering
  \includegraphics[width=\linewidth]{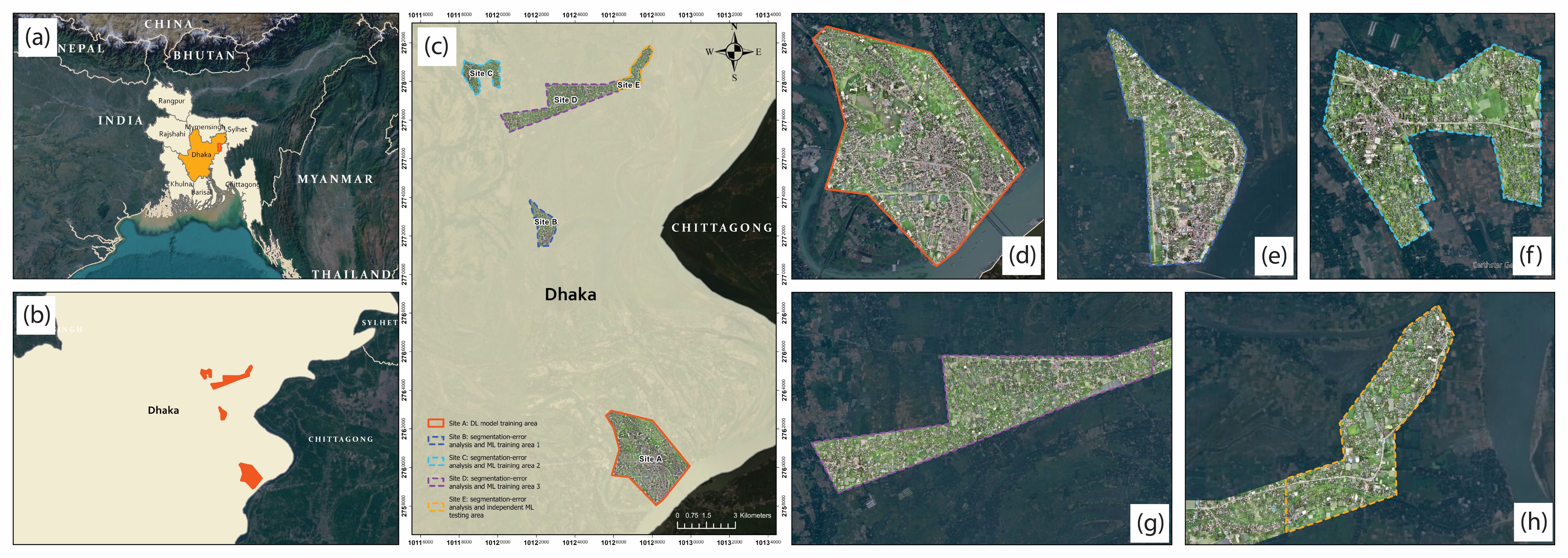}
  \caption{Location and analytical roles of the five UAV study sites in Dhaka Division, Bangladesh: (a-b) national and divisional context; (c) site distribution; and (d-h) orthophoto extents of Sites A-E respectively.}
  \label{fig:fig2}
\end{figure}

We surveyed five distinct locations (designated as Sites A through E) in the Dhaka Division of Bangladesh utilizing unmanned aerial vehicles (UAVs). These sites demonstrated compact and irregular settlement patterns characteristic of peri-urban and semi-rural areas, as illustrated in (\autoref{fig:fig2}). The sites exhibited variations in roof materials, building density, vegetation interference, shadowing, and boundary complexity.
We created reference building footprint polygons only for Site A, which were used for DL training and validation. Sites B-D contributed to footprint-level ML development, and Site E was reserved for independent testing. Our collected 8-bit RGB orthophoto has a spatial resolution of 15 cm. We used WGS 1984 UTM Zone 46N as a projected coordinate system. The orthophoto and corresponding reference building polygons of Site A were converted into paired image chips and pixel-wise label masks to construct the semantic-segmentation training dataset. By employing classified tiles metadata, we created 17,512 image-mask sample pairs using 256 x 256-pixel chips and a 128 x 128-pixel stride (Figure S1). Ten percent of these samples were set aside for validation.

\subsection{Deep-learning-based building-footprint extraction}
\begin{figure}[htbp]
  \centering
  \includegraphics[width=\linewidth]{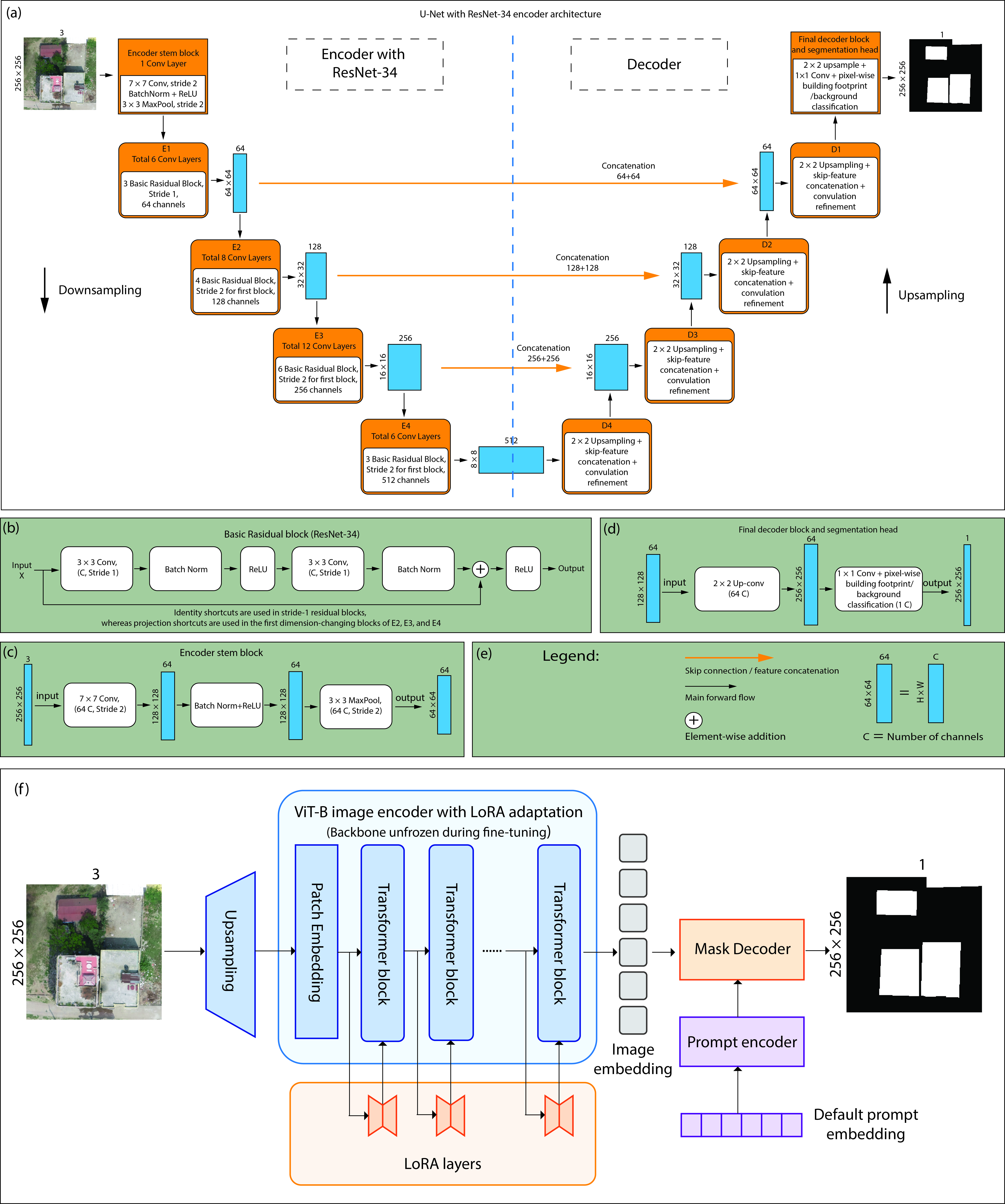}
  \caption{Part (a-e) U-Net architecture in with a ResNet-34 encoder used to generate 256 x 256-pixel binary building masks and Part (f) SAM-LoRA architecture with a ViT-B image encoder, LoRA adaptation, default prompt embedding, and mask decoder.}
  \label{fig:fig3}
\end{figure}

We approached building extraction as a binary semantic segmentation task, utilizing two distinct architectures (\autoref{fig:fig3}). The convolutional U-Net architecture incorporated an ImageNet-pretrained ResNet-34 encoder, which featured multiscale skip connections and a symmetric decoder to accurately delineate fine boundaries and small structures (He et al., 2016; Ronneberger et al., 2015). Conversely, SAM-LoRA modified the pretrained Segment Anything Model by integrating low-rank modules into the ViT-B image encoder, facilitating task-specific fine-tuning while preserving the pretrained features (Hu et al., 2022; Kirillov et al., 2023). In the ArcGIS implementation, which did not require prompts, default prompt embeddings and a mask decoder were employed to generate binary building masks.
Both segmentation models were trained on the same Site A image–mask dataset for 50 epochs with a batch size of 4 and the validation loss was assessed following each epoch to monitor the training process. SAM-LoRA was optimized using a learning rate of $1.0 \times 10^{-4}$, whereas U-Net employed discriminative learning rates ranging from  $1.0 \times 10^{-5}$ to  $1.0 \times 10^{-4}$ Neither early stopping nor class balancing techniques were employed. We adopted a standardized training protocol to reduce bias from training conditions and to produce candidate footprint populations from two architecturally diverse segmentation sources. The deep learning phase was not designed as a comprehensive architecture-level benchmark; selecting only the apparently superior model would have restricted the candidate footprint database and introduced bias in the subsequent consolidation and quality assessment phases. Each model was independently applied to Sites B-E, resulting in eight segmentation masks.

\subsection{Mask-to-vector conversion and footprint geometric regularization}

\begin{figure}[htbp]
  \centering
  \includegraphics[width=\linewidth]{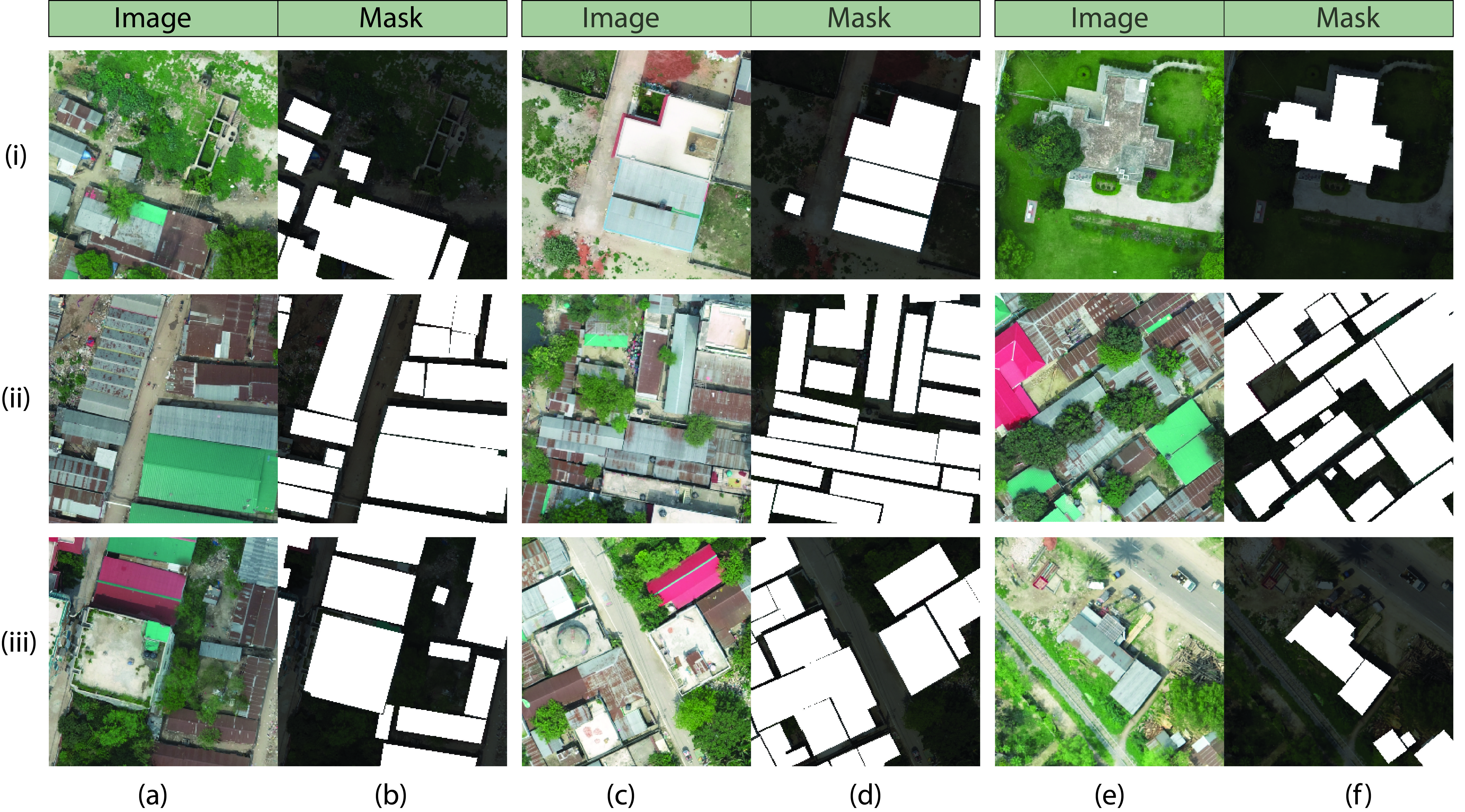}
  \caption{Representative RGB image chips and corresponding binary building-label tiles used for DL model training in Site A.}
  \label{fig:fig4}
\end{figure}

For each model–site combination, the connected foreground regions identified in the binary segmentation masks were delineated and transformed into vector polygon geometries (\autoref{fig:fig4}). These polygons were then refined using right-angle geometric regularization with a 2 m tolerance, which served to minimize boundary irregularities and enhance the geometric consistency of the building footprints. This standard post-processing procedure generated GIS-compatible candidate footprints by smoothing stair-step edges, eliminating extraneous vertices, and correcting local boundary distortions resulting from raster-to-vector conversion and building-footprint polygonization (Girard et al., 2021; Sanca et al., 2023; Zorzi et al., 2022). We applied identical settings to both U-Net and SAM-LoRA outputs to ensure that object-level comparisons remained unaffected by variations in vector-processing conditions. Regularization was regarded as a geometric post-processing step rather than evidence of semantic accuracy or database readiness. Although it enhances cartographic form, it does not verify whether a polygon accurately represents a building, rectify merged or fragmented objects, recover missing components, or determine suitability for database integration. Consequently, we treated the regularized polygons as intermediate candidate objects pending further object-level quality assessments. In this study, the enhancement of database quality pertains to the subsequent ML-based screening and retention phase, rather than solely to geometric regularization (Niroshan and Carswell, 2025; Pepe et al., 2021).

\subsection{Object-level multidomain geospatial feature engineering}

In the feature-engineering phase, we converted each regularized footprint into an object-level representation. Instead of merely considering the segmentation result as a raster mask, each polygon was characterized by its internal geometry, the spatial context around it, and the roof appearance derived from the raster. We used an error-driven, domain-informed approach to select 24 descriptors that address common post-segmentation issues, such as irregular boundaries, distorted or elongated polygons, merged and fragmented objects, isolated false positives, unusual local settlement patterns, and spectral or textural inconsistencies with building roofs (\autoref{tab:tab1}). Features were calculated on the complete footprint layers specific to each model before consolidating footprints across models, adhering to the spatial-exclusivity constraint, and performing merging, partitioning, and class balancing to maintain the neighborhood structure necessary for contextual variables. The site identifier, source model, and manual quality class were kept as metadata and not included in model inputs.

\begin{table}[htbp]
  \footnotesize
  \centering
  \renewcommand{\arraystretch}{1} 
  \caption{Mathematical definition and diagnostic interpretation of the geometric, spatial-contextual, and raster-derived descriptors (also see Figure A.1-3).}
  \label{tab:tab1}
  \begin{tabular}{c p{7.5cm} p{6.5cm}}
    \toprule
    \textbf{Code} & \textbf{Descriptor with Equation / Formulation} & \textbf{Diagnostic Interpretation} \\
    \midrule
    \multicolumn{3}{l}{\textbf{1. Geometric Descriptors (12 Predictors)}} \\
    \midrule
    (a) & Footprint area, $A_i = \operatorname{area}(\Omega_i)$ &  Planar extent of candidate footprint. \\
    (b) & Edge-length variance, $\mathrm{E_V,}_i=\frac{1}{n_{e,i}}\sum_{r=1}^{n_{e,i}}(e_{ir}-\bar e_i)^2$ \newline $\bar e_i=\frac{1}{n_{e,i}}\sum_{r=1}^{n_{e,i}}e_{ir}$
 & Quantifies variation among boundary-segment lengths. \\
    (c) & Compactness index, $\mathrm{C}_i=\dfrac{4\pi A_i}{P_i^2}$ & Shape compactness relative to area and perimeter. \\
    (d) & Vertex density, $\mathrm{V_D,}_i=\dfrac{N_{v,i}}{P_i}$ & Ioundary segmentation intensity and vertex concentration. \\
    
    (e) & Solidity (convex fill), $\mathrm{S}_i=\dfrac{A_i}{A_{\mathrm{convex},i}}$ & Convex-hull occupancy and sensitivity to concavities. \\
    (f) & Near-right-angle ratio,  \newline$\mathrm{R\_RA}_i=\dfrac{1}{n_{\alpha,i}}\sum_{r=1}^{n_{\alpha,i}}
\mathbf{1}\left(|\alpha_{ir}-90^\circ|\le15^\circ\right)$ \newline
$\alpha_{ir}=\cos^{-1}\!\left(\dfrac{\mathbf{v}_{ir,1}\cdot\mathbf{v}_{ir,2}}
{\left\Vert\mathbf{v}_{ir,1}\right\Vert\left\Vert\mathbf{v}_{ir,2}\right\Vert}\right)$ & Proportion of vertices satisfying right-angle tolerance.\\
    (g) & Principal-axis elongation, $\mathrm{PCA\_Elong}_i=\mathrm{Major}_i/\mathrm{Minor}_i$ if $\lambda_2>10^{-12}$, else $0$ \newline $\lambda_{1,2}=\dfrac{s_{xx}+s_{yy}\pm\sqrt{(s_{xx}-s_{yy})^2+4s_{xy}^2}}{2}$ \newline
$\mathrm{Major}_i=\sqrt{12\lambda_1}$, $\mathrm{Minor}_i=\sqrt{12\lambda_2}$ 
 & Elongation from the ratio of principal-axis lengths. \\
 
    (h) & Orthogonality score, $\mathrm{S\_ortho}_i=\frac{1}{n_{e,i}}\sum_{r=1}^{n_{e,i}}\mathbf{1}(d_{ir}\le10^\circ)$ \newline
$a_{ir}=\left[\frac{180}{\pi}\operatorname{atan2}(\Delta y_{ir},\Delta x_{ir})\right]\bmod 180^\circ$ \newline
$\mu_i=\left[\frac{1}{2}\operatorname{atan2}(\overline{\sin 2a_{ir}},\overline{\cos 2a_{ir}})\right]\bmod 180^\circ$ \newline
$d_{ir}=\min\{\Delta_{180}(a_{ir},\mu_i),\,|\Delta_{180}(a_{ir},\mu_i)-90^\circ|\}$ & Alignment of boundary edges with dominant orthogonal axes. \\
    (i) & Principal-axis orientation, \newline$\mathrm{PCA\_Orient}_i=\left[\frac{180}{\pi}\frac{1}{2}\operatorname{atan2}(2s_{xy},s_{xx}-s_{yy})\right]\bmod 180^\circ$ & Dominant principal-axis direction. \\
    (j) & MBR occupancy ratio, $\mathrm{MBR}_i=\dfrac{A_i}{A_{\mathrm{MBR},i}}$ & Occupancy of oriented minimum bounding rectangle. \\
    (k) & Mean edge length, $\mathrm{Edge\_M}_i=\dfrac{1}{n_{e,i}}\sum_{r=1}^{n_{e,i}}e_{ir}$ & Average boundary-segment length. \\
    (l) & Perimeter-area fractal dimension, $D\mathrm{PA}_i=\dfrac{2\ln(P_i/4)}{\ln(A_i)}$ & Boundary complexity from perimeter-area scaling. \\
    \midrule
    \multicolumn{3}{l}{\textbf{2. Spatial-Contextual Descriptors (6 Predictors)}} \\
    \midrule
    (a) & Neighbor count (50\,m), $\mathrm{C}_i=\mathrm{Join\_Count}_i-1=|N_{50}(i)|$ & Local footprint density around focal object. \\
    (b) & Orientation dispersion,
$\mathrm{Orient\_D}_i=1-R_i$ \newline $\theta_j=\operatorname{atan2}(\Delta y_{\mathrm{longest},j},\Delta x_{\mathrm{longest},j})$ \newline
$R_i=\dfrac{\sqrt{\left(\sum_{j\in N_{15}(i)}\sin\theta_j\right)^2+\left(\sum_{j\in N_{15}(i)}\cos\theta_j\right)^2}}{m_i}$  & Directional heterogeneity among neighbor footprints. \\
    (c) & Local Moran's $I$ (Area),\newline
$\mathrm{LM\_I}(i)=\dfrac{x_i-\bar x}{\operatorname{Var}_N(x)}\sum_{j\in N_{15}(i)}w_{ij}(x_j-\bar x)$  & Quantifies local spatial association in footprint area. \\
    (d) & Cluster size, $\mathrm{CL}_{\mathrm{SIZE}} = |C_i|$ & Number of objects in focal proximity cluster. \\
    (e) & Nearest-neighbor dist., $d_{\mathrm{NN}} = \min(d_{ij})$ & Isolation from closest neighboring footprint. \\
    (f) & Local area z-score,  $Z_{\mathrm{area}} = \frac{A_i - \mu_{\mathrm{local}}}{\sigma_{\mathrm{local}}}$ & Footprint area deviation relative to local neighborhood. \\
    \midrule
    \multicolumn{3}{l}{\textbf{3. Spectral and Raster-Derived Descriptors (6 Predictors)}} \\
    \midrule
    (a) & Norm. red proportion, $r = \frac{\bar R}{\bar R+\bar G+\bar B}$ & Red contribution relative to total RGB intensity. \\
    (b) & Norm. green proportion, $g = \frac{\bar G}{\bar R+\bar G+\bar B}$ & Green contribution relative to total RGB intensity. \\
    (c) & Red-band std. dev., $\sigma_R = \sqrt{\frac{1}{n}\sum_{k=1}^{n} (R_i - \bar{R})^2}$ & Within-footprint red-band heterogeneity. \\
    (d) & Mean $\mathrm{PC}_1$ score,  $\bar{\mathrm{PC}_1} = \frac{1}{n}\sum_{k=1}^{n} {PC1_k}$ & Dominant correlated RGB response in footprint. \\
    (e) & Mean local gray std. dev.,  $\bar{\sigma}_{\mathrm{gray}} = \frac{1}{n} \sum_{k=1}^{n} {\sigma}_{3\times3}\frac{R+G+B}{3}$ & Average local grayscale variability. \\
    (f) & Mean local gray range, $\bar{R}_{\mathrm{gray}} = \frac{1}{n} \sum_{k=1}^{n} {R}_{3\times3}\frac{R+G+B}{3}$ & Average local contrast as an edge-strength proxy. \\
    \bottomrule
  \end{tabular}
\end{table}

\subsubsection{Geometric features}
Geometric characteristics define the internal structure and boundary outline of each footprint after processing. These characteristics are essential for identifying irregular polygon shapes, overly intricate boundaries, insufficient rectangularity, uneven vertex distribution, and low compactness, which are frequently observed in building footprints that are fragmented, merged, distorted, or not adequately regularized (\autoref{tab:tab1}).

\subsubsection{Spatial-contextual features}

Spatial-contextual features assess each footprint by examining its relationship with nearby building patterns and local settlement structures. To identify characteristics such as isolation, local density, cluster membership, orientation consistency, spatial association in footprint area, and area anomaly, several metrics were applied: nearest-neighbor distance, the number of neighbors within 50 meters, the size of proximity-connected clusters, the dispersion of neighborhood orientation, Local Moran's I, and the local area z-score (\autoref{tab:tab1}). The implementation of Local Moran's I was based on the local-indicator framework developed by (Anselin, 1995).

\subsubsection{Raster-derived spectral and texture features}
Raster-derived features encapsulate the spectral and textural attributes of UAV image pixels within each refined footprint (\autoref{tab:tab1}). This domain offers image-content evidence that augments geometric and contextual information, particularly in distinguishing false-positive polygons that may geometrically mimic buildings but actually encompass vegetation, roads, shadows, bare soil, or other non-building surfaces. The normalized proportions of red and green, the standard deviation of the red band, and the mean score of the first principal component delineate the color, brightness, and spectral heterogeneity within footprints. In contrast, the mean local grayscale standard deviation and the mean local grayscale range define texture variation and local contrast, serving as a proxy for edge strength (Table 4) (Blaschke, 2010; Du et al., 2015; Jolliffe and Cadima, 2016; Ma et al., 2017).\\
In \autoref{tab:tab1}, A and P denote footprint area and perimeter; $A_convex$ and AMBR denote convex-hull and minimum-bounding-rectangle areas; $\lambda_1$ and $\lambda_2$ are the eigenvalues of the vertex-coordinate covariance matrix; $\theta$ denotes dominant orientation; n denotes the relevant number of edges, neighbors, or pixels; and $N_i$ denotes the neighborhood of footprint i. The descriptors were grouped into geometric (G), spatial-contextual (C), and raster-derived (R) domains.

\subsection{Object-level reference dataset curation and spatially independent partitioning}
\begin{figure}[htbp]
  \centering
  \includegraphics[width=0.50\linewidth]{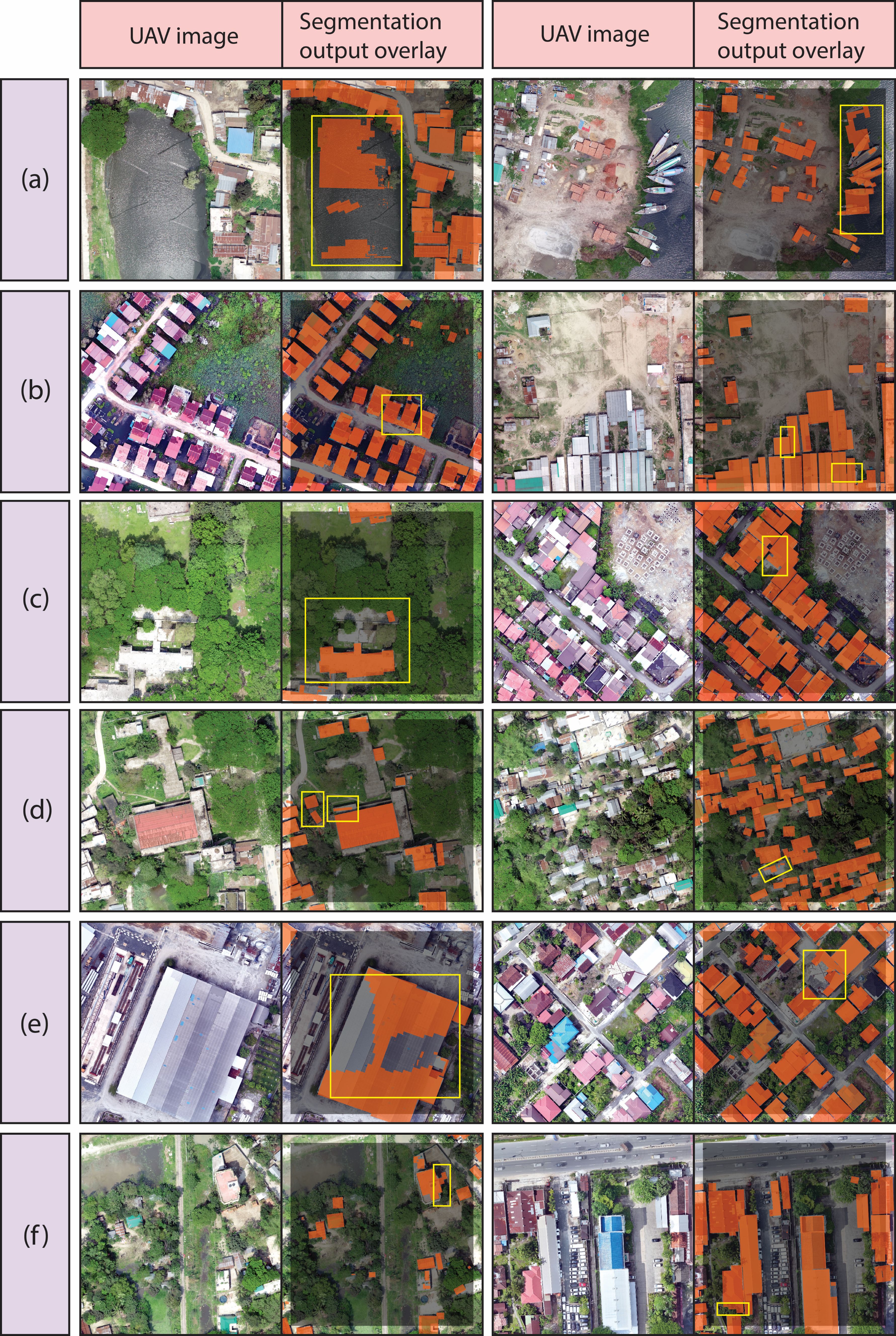}
  \caption{Representative segmentation-derived footprint errors used for manual reference labelling: (a) false-positive building predictions; (b) merged objects or under-segmentation; (c) geometric distortion; (d) split objects or over-segmentation; (e) incomplete boundary coverage; and (f) zigzag or noisy boundaries. Yellow boxes identify representative errors.}
  \label{fig:fig5}
\end{figure}

Following feature engineering, each footprint from both models was manually evaluated against the corresponding 15-cm orthophoto and categorized as either acceptable (class 0) or erroneous (class 1). A footprint was deemed acceptable if its position, completeness, boundary alignment, and separation from nearby structures appeared visually credible. Erroneous objects included false-positive building predictions, footprints that were merged or insufficiently segmented, geometric distortions, footprints that were split or excessively segmented, incomplete boundary coverage, and boundaries that were zigzagged or noisy (\autoref{fig:fig5}). \\
Subsequently, a consolidation of cross-model footprints was executed under a spatial-exclusivity constraint to produce a single candidate footprint layer for each site (Table A.1). Instances where U-Net and SAM-LoRA polygons overlapped or were nearly coincident were considered as competing depictions of the same visible building and were resolved using the source-neutral retention hierarchy. The rule focused on database suitability over model preference: valid candidates were chosen over incorrect ones; candidates sharing the same label were resolved based on boundary alignment, completeness, separation from adjacent buildings, and geometric plausibility; and detections from a single model were retained and labeled through orthophoto interpretation. This process minimized duplicate observations across models while ensuring that each competing building instance contributed only one representative candidate to the consolidated analytical layer. \\
The merged layers from Sites B-D formed the model-development partition. We applied controlled class balancing only to this partition. All 3,713 erroneous objects remaining after consolidation were retained, whereas 4,985 acceptable objects were selected through spatially controlled under sampling, yielding 8,698 development samples without synthetic resampling or recalculation of contextual features (He and Garcia, 2009). Site E underwent the same feature computation, manual reference labelling, cross-model footprint consolidation under the spatial-exclusivity constraint, and merged-layer preparation, but it was not subjected to class-ratio adjustment. All 4,162 Site E objects (3,025 acceptable and 1,137 erroneous) were retained as an independent feature dataset for testing. Site E contributed neither to model fitting nor to hyperparameter optimization and class balancing (Roberts et al., 2017).

\subsection{Machine-learning model development and footprint quality classification}
\begin{figure}[htbp]
  \centering
  \includegraphics[width=\linewidth]{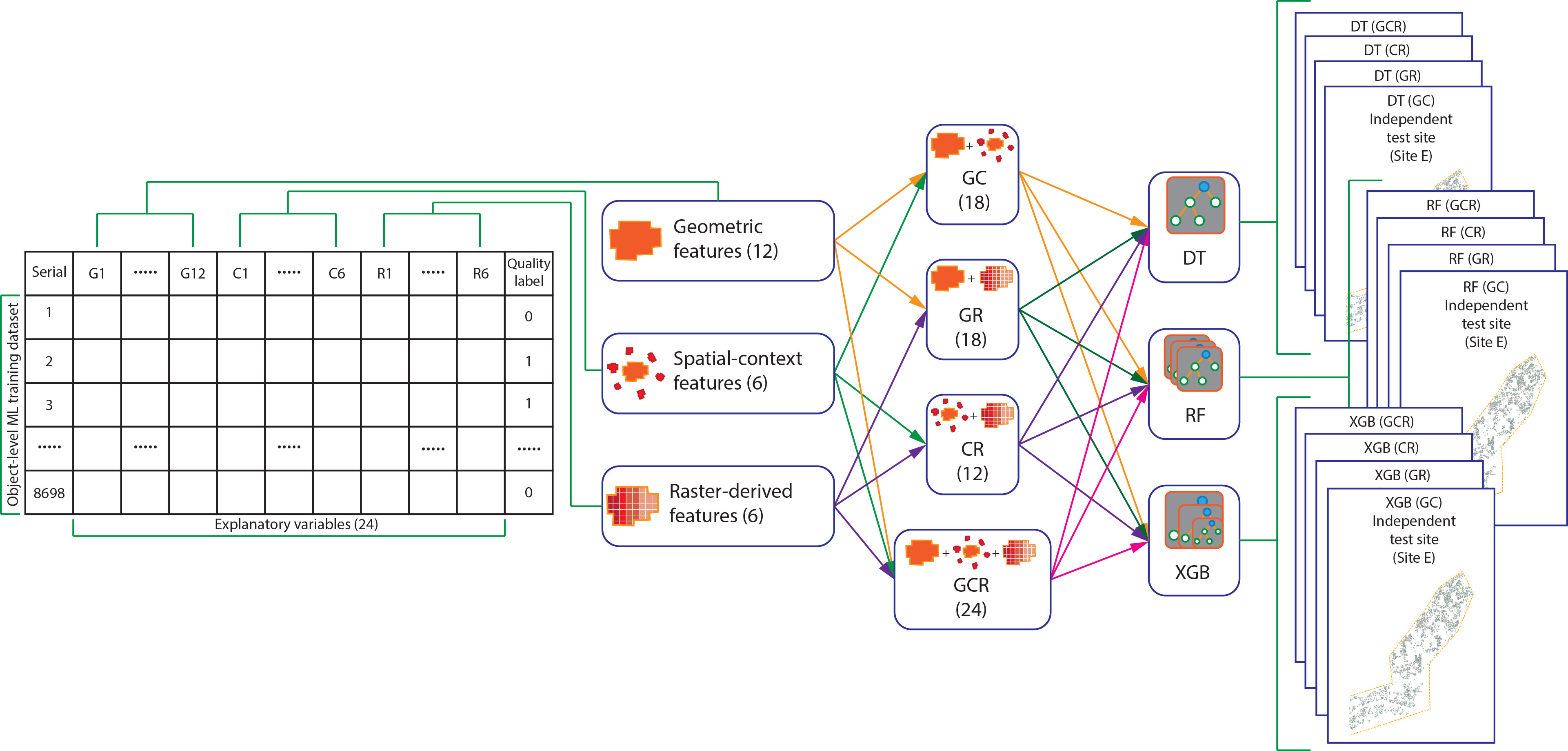}
  \caption{Experimental design for ML footprint-quality classification. Geometric (G), spatial-contextual (C), and raster-derived (R) predictors formed four feature configurations for Decision Tree, Random Forest, and XGBoost, yielding 12 trained configurations.}
  \label{fig:fig6}
\end{figure}

The quality of footprints was assessed through a supervised binary classification approach, utilizing a consolidated and class-balanced dataset from Sites B-D for model development (\autoref{fig:fig6}). Objects were categorized into two classes: acceptable (class 0) and erroneous (class 1), with the latter being the positive class due to the goal of pinpointing footprints that needed removal or further examination in the consolidated database. A total of 24 predictors served as explanatory variables, while site, source model, and reference label were omitted from the model inputs. \\
Within a standardized automated machine-learning framework, classifiers such as Decision Tree (DT, Random Forest (RF), and Extreme Gradient Boosting (XGB) were trained and assessed. Each feature configuration underwent evaluation through uniform training and validation methods (Breiman, 2001; Breiman et al., 1984; Chen and Guestrin, 2016). A 10 percent validation percentage was applied, and the demographic parity ratio served as the fairness metric across all machine learning models. Each learning algorithm was trained with four different predictor-domain setups: GC (18 variables), GR (18), CR (12), and GCR (24). These 12 model-feature configurations utilized the same development samples and class definitions (\autoref{fig:fig6}). Model fitting and configuration selection were restricted to the model-development partition from Sites B-D. Each of the 12 locked configurations of the classifier underwent evaluation using the Site E dataset, which was spatially independent. Site E was deliberately excluded from processes such as model fitting, internal validation, feature-configuration selection, hyperparameter tuning, class-ratio adjustment, and decision-threshold modification.

\subsection{Independent test evaluation and data-quality impact assessment}
The independent test's performance was assessed using various metrics, including accuracy, precision, recall (sensitivity), specificity, F1-score, balanced accuracy, the Matthews correlation coefficient (MCC), and Cohen's kappa. Due to the imbalance in test classes and the distinct operational implications of the two error types, interpretation was not solely based on accuracy. Recall was used to indicate the detection of erroneous footprints, precision measured the reliability of flagged objects, specificity indicated the preservation of acceptable footprints, and balanced accuracy along with MCC provided a summary of class-balanced discrimination (Chicco and Jurman, 2020; Cohen, 1960; Sokolova and Lapalme, 2009).
The classification performance was evaluated using standard metrics derived from the binary confusion matrix. Let TP, TN, FP, and FN denote true positives, true negatives, false positives, and false negatives, respectively. The total number of samples evaluated was defined as

\begin{equation}
N = TP + TN + FP + FN
\end{equation}

The model-level performance indicators were computed as follows:
\begin{equation}
\begin{aligned}
\mathrm{Accuracy} &=
\frac{TP + TN}{N}, \\[4pt]
\mathrm{Precision} &=
\frac{TP}{TP + FP}, \\[4pt]
\mathrm{Recall} &=
\frac{TP}{TP + FN}, \\[4pt]
\mathrm{Specificity} &=
\frac{TN}{TN + FP}, \\[4pt]
F_1 &=
\frac{2 \times \mathrm{Precision} \times \mathrm{Recall}}
{\mathrm{Precision} + \mathrm{Recall}}, \\[4pt]
\mathrm{Balanced\ Accuracy} &=
\frac{\mathrm{Recall} + \mathrm{Specificity}}{2}.
\end{aligned}
\end{equation}

To account for the reliability of classification under class imbalance, the Matthews correlation coefficient (MCC) was calculated as
\begin{equation}
\mathrm{MCC} = \frac{TP \times TN - FP \times FN}{\sqrt{(TP + FP)(TP + FN)(TN + FP)(TN + FN)}}
\end{equation}

Cohen's kappa coefficient was calculated as
\begin{equation}
\begin{aligned}
\kappa &= \frac{p_o - p_e}{1 - p_e}, \\
p_o &= \frac{TP + TN}{N}, \\
p_e &= \frac{
(TP + FP)(TP + FN)
+
(TN + FN)(TN + FP)
}{N^2}.
\end{aligned}
\end{equation}

In addition to model-level evaluation, the database-level effect of ML-based screening was quantified from the same confusion matrix. Predicted erroneous footprints were considered as objects flagged for removal or manual review, whereas predicted acceptable footprints were considered as retained objects. Accordingly, the database-level quantities were defined as

\begin{equation}
\begin{aligned}
F_{\mathrm{flagged}}  &= TP + FP, \\
F_{\mathrm{retained}} &= TN + FN, \\
E_{\mathrm{residual}} &= FN, \\
A_{\mathrm{retained}} &= TN.
\end{aligned}
\end{equation}

The initial error proportion, residual error proportion after screening, retained-data purity, and corresponding error reductions were calculated as
\begin{equation}
\begin{aligned}
e_0 &= \frac{TP + FN}{N}, \\
e_1 &= \frac{FN}{TN + FN}, \\
P_{\mathrm{retained}} &= \frac{TN}{TN + FN}, \\
\Delta e &= e_0 - e_1, \\
R_{\mathrm{error}} &= 100 \times \frac{\Delta e}{e_0}.
\end{aligned}
\end{equation}

The preferred configuration was identified by jointly considering independent-test discrimination, residual error burden, and loss of acceptable footprints rather than any single metric.

\section{Results}
Results are presented in two phases. Initially, concise DL training diagnostics are offered, utilizing only the final-epoch validation metrics to ensure adequate convergence for the two segmentation-based footprint generators. The primary analysis then examines the 12 fixed classifier-feature configurations applied to the spatially independent consolidated Site E dataset, which comprised 4,162 footprint objects, including 1,137 erroneous and 3,025 acceptable footprints. Consequently, prior to ML screening, the independent test database consisted of 27.32\% erroneous footprints and a purity of 72.68\% for acceptable footprints.

\subsection{Deep-learning training diagnostics}
The purpose of the training diagnostics was to ensure that the segmentation models could generate viable candidate footprint populations for further object-level analysis, rather than to compare the two DL architectures. To prevent confusion between the stage with the lowest validation loss and the final model state, \autoref{tab:tab2} exclusively presents the validation metrics from the 50$^{th}$ epoch. At this concluding epoch, both SAM-LoRA and U-Net demonstrated similar performance in identifying foreground building classes and generated sufficiently dependable segmentation-derived footprint candidates for the subsequent stages of consolidation and quality control.

\begin{table}[htbp]
\centering
\caption{Final-epoch ($50^{th}$) deep-learning training diagnostics for the two segmentation-based footprint generators.}
\label{tab:tab2}
\resizebox{\linewidth}{!}{%
\begin{tabular}{llccccccc}
\hline
\textbf{Model} & \textbf{Backbone} & \textbf{Val. loss} & \textbf{Accuracy} & \textbf{Dice} & \textbf{Precision} & \textbf{Recall} & \textbf{F1-score} \\
\hline
SAM-LoRA & ViT-B & 0.1336 & 0.9439 & 0.8855 & 0.9164 & 0.9111 & 0.9138 \\
U-Net & ResNet-34 & 0.2387 & 0.9437 & 0.8720 & 0.9193 & 0.9077 & 0.9134 \\
\hline
\end{tabular}%
}
\end{table}

\subsection{Independent-test classification performance}
Performance on independent tests varied significantly depending on both the classifier type and the feature-domain setup (\autoref{tab:tab3}). The most effective configuration overall was the DT, which was trained using a combination of geometric and spatial-contextual predictors. This model achieved an accuracy of 95.31\%, precision of 95.11\%, recall of 87.34\%, specificity of 98.31\%, and an F1-score of 91.06\%. Its balanced accuracy, MCC, and Cohen's kappa were 92.82\%, 0.880, and 0.879, respectively, demonstrating strong class-balanced discrimination on the unseen site. DT models consistently showed the highest discrimination in independent tests. The GR setup yielded a similar accuracy of 95.22\% and an F1-score of 90.86\%, although its recall and MCC were slightly lower than those of the GC setup. XGB also performed well when geometric and contextual predictors were combined, while the CR setup resulted in the weakest outcomes for all three learners. This trend suggests that raster-derived and neighborhood variables alone were inadequate to substitute direct footprint-shape information for detecting object-level errors.

\begin{table}[htbp]
\centering
\caption{Independent-test classification performance of the 12 model-feature configurations on Site E.}
\label{tab:tab3}
\resizebox{\linewidth}{!}{%
\begin{tabular}{llcccccccc}
\hline
\textbf{Model} & \textbf{Feature group} & \textbf{Acc.} & \textbf{Prec.} & \textbf{Rec.} & \textbf{Spec.} & \textbf{F1} & \textbf{BA} & \textbf{MCC} & $\mathbf{\kappa}$ \\
\hline
XGB & GCR & 89.48 & 76.10 & 89.62 & 89.42 & 82.31 & 89.52 & 0.754 & 0.749 \\
 & GR & 88.27 & 73.33 & 89.71 & 87.74 & 80.70 & 88.72 & 0.732 & 0.724 \\
 & CR & 81.52 & 61.90 & 84.17 & 80.53 & 71.34 & 82.35 & 0.597 & 0.582 \\
 & GC & 91.33 & 81.14 & 88.92 & 92.23 & 84.85 & 90.57 & 0.790 & 0.788 \\
\hline
RF & GCR & 87.63 & 71.84 & 89.97 & 86.74 & 79.89 & 88.36 & 0.721 & 0.711 \\
 & GR & 82.56 & 62.28 & 91.64 & 79.14 & 74.16 & 85.39 & 0.643 & 0.617 \\
 & CR & 83.09 & 64.54 & 84.52 & 82.55 & 73.19 & 83.53 & 0.623 & 0.612 \\
 & GC & 90.80 & 79.41 & 89.53 & 91.27 & 84.17 & 90.40 & 0.780 & 0.777 \\
\hline
DT & GCR & 94.47 & 92.11 & 87.25 & 97.19 & 89.61 & 92.22 & 0.859 & 0.859 \\
 & GR & 95.22 & 95.10 & 86.98 & 98.31 & 90.86 & 92.65 & 0.878 & 0.876 \\
 & CR & 87.27 & 86.70 & 63.06 & 96.36 & 73.01 & 79.71 & 0.664 & 0.650 \\
 & GC & 95.31 & 95.11 & 87.34 & 98.31 & 91.06 & 92.82 & 0.880 & 0.879 \\
\hline
\multicolumn{10}{l}{\small \textit{Note:} Acc. = accuracy; Prec. = precision; Rec. = recall/sensitivity; Spec. = specificity; BA = balanced accuracy; $\kappa$ = Cohen's kappa.} \\
\multicolumn{10}{l}{\small Values except MCC and $\kappa$ are percentages.}
\end{tabular}%
}
\end{table}

\begin{figure}[htbp]
  \centering
  \includegraphics[width=0.55\linewidth]{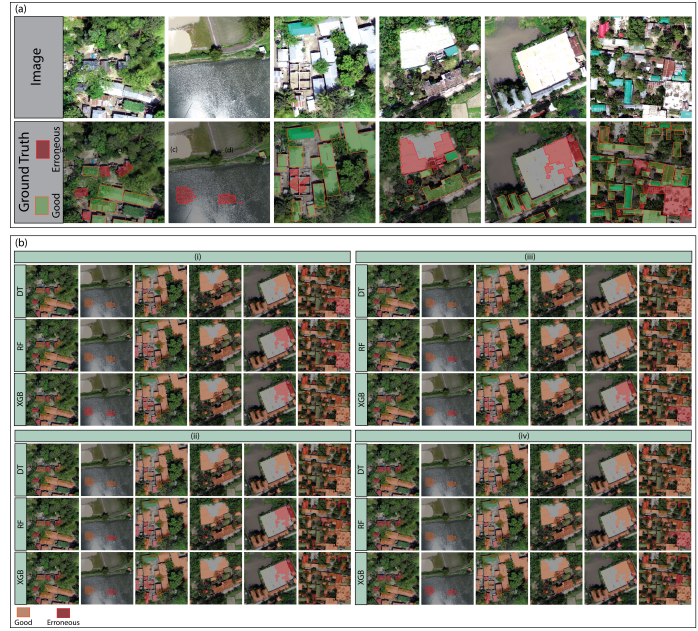}
  \caption{Qualitative Site E footprint-quality classification outputs for four feature-domain configurations. Part (a) displays six representative image subsets alongside their corresponding ground-truth manual labels, where green denotes good/acceptable footprints and red denotes erroneous ones labeled by human experts. Part (b) shows the prediction outputs from Decision Tree (DT), Random Forest (RF), and XGBoost (XGB) classifiers across the four configurations: (i) Geometry and Context - GC, (ii) Geometry and Raster - GR, (iii) Context and Raster - CR, and (iv) Geometry, Context, and Raster - GCR. In the classifier prediction panels (b), orange outlines/fills denote acceptable footprints, while red denotes erroneous footprints classified by ML models.}
  \label{fig:fig7}
\end{figure}

A qualitative examination of the windows at representative Site E offers a visual complement to the metrics from independent tests (\autoref{fig:fig7}). To maintain the connection between closely related feature-domain outputs, the figure is divided into two consecutive sections rather than separate visual references. The outputs from the GC and GR configurations demonstrated a more reliable distinction between correct and incorrect footprints across various settlement environments. In contrast, the CR setup exhibited noticeably weaker differentiation in areas that are dense, vegetated, and spectrally diverse. These instances also clarify why the complete 24-feature setup did not necessarily surpass the geometry-inclusive combinations: redundant spectral-contextual information did not consistently enhance the screening of boundary errors at the object level once the primary geometric signal was captured.

\subsection{Effect of feature-domain composition}
Feature-domain aggregation was applied to evaluate how predefined descriptor combinations contribute to discrimination, rather than ranking individual domains within a framework that normalizes feature counts (\autoref{tab:tab4}). Across the three classifiers, the GC setup consistently achieved the highest scores: 92.48\% for accuracy, 86.69\% for the F1-score, 91.27\% for balanced accuracy, 0.817 for MCC, and 0.815 for Cohen's kappa. Its specificity of 93.94\% indicates that this setup effectively reduced false positives while maintaining the integrity of valid footprint objects.\\
The differing number of descriptors across domains require careful consideration, particularly for the CR setup. However, the trends observed were not solely influenced by the number of descriptors: the comprehensive 24-variable setup did not yield the best average performance, and the two 18-variable setups, GC and GR, showed varying levels of discrimination in independent tests. These findings suggest that within the predefined object-level descriptor set, GC was the most effective feature-domain composition, while evidence derived solely from contextual and raster sources did not match the diagnostic value of direct footprint-shape descriptors.

\begin{table}[htbp]
\centering
\caption{Group-wise average independent-test impact of the feature-domain configurations.}
\label{tab:tab4}
\resizebox{\linewidth}{!}{%
\begin{tabular}{lcccccccccc}
\hline
\textbf{Feature group} & \textbf{Acc.} & \textbf{Prec.} & \textbf{Rec.} & \textbf{Spec.} & \textbf{F1} & \textbf{BA} & \textbf{MCC} & \textbf{Residual error} & \textbf{Purity} & \textbf{Relative reduction} \\
\hline
GCR & 90.53 & 80.02 & 88.95 & 91.12 & 83.94 & 90.03 & 0.778 & 4.35 & 95.65 & 84.08 \\
GR & 88.68 & 76.90 & 89.45 & 88.40 & 81.91 & 88.92 & 0.751 & 4.26 & 95.74 & 84.41 \\
CR & 83.96 & 71.05 & 77.25 & 86.48 & 72.51 & 81.86 & 0.628 & 8.69 & 91.31 & 68.20 \\
GC & 92.48 & 85.22 & 88.60 & 93.94 & 86.69 & 91.27 & 0.817 & 4.36 & 95.64 & 84.05 \\
\hline
\multicolumn{11}{l}{\small \textit{Note:} Residual error, purity, and relative reduction refer to the retained footprint database after ML-based screening.}
\end{tabular}%
}
\end{table}

\subsection{Database-level footprint-quality impact}
The chosen DT (GC) setup identified 1044 footprints as incorrect, with 993 being true positives and 51 false positives. It maintained 3118 footprints, which included 2974 true negatives and 144 false negatives. As a result, the initial error rate decreased from 27.32\% to a remaining error of 4.62\% [FN/(TN + FN)], while the purity of the retained data improved from 72.68\% to 95.38\% [TN/(TN + FN)]. These confusion-matrix figures also account for the 95.11\% precision of the flagged set [TP/(TP + FP)] and the 98.31\% preservation of acceptable footprints [TN/(TN + FP)]. This translates to an absolute error reduction of 22.70 percentage points and a relative reduction of 83.09\% in the retained Site E footprint database. This outcome is significant operationally because the model eliminated 87.34\% of erroneous footprints while retaining 98.31\% of acceptable ones. Although RF (GR) achieved the highest recall (91.64\%) and the lowest residual error rate (3.82\%), it also resulted in more false positive removals. Therefore, the preferred configuration provided a more favorable balance between reducing errors and preserving footprints (\autoref{tab:tab5}).

\begin{table}[htbp]
\centering
\caption{Operational database-quality impact for selected high-performing configurations on Site E.}
\label{tab:tab5}
\resizebox{\linewidth}{!}{%
\begin{tabular}{lccccccccc}
\hline
\textbf{Configuration} & \textbf{Flagged} & \textbf{TP} & \textbf{FP} & \textbf{Retained} & \textbf{FN} & \textbf{TN} & \textbf{Residual error} & \textbf{Purity} & \textbf{Relative reduction} \\
\hline
DT (GC) & 1044 & 993 & 51 & 3118 & 144 & 2974 & 4.62 & 95.38 & 83.09 \\
DT (GR) & 1040 & 989 & 51 & 3122 & 148 & 2974 & 4.74 & 95.26 & 82.65 \\
DT (GCR) & 1077 & 992 & 85 & 3085 & 145 & 2940 & 4.70 & 95.30 & 82.80 \\
XGB (GC) & 1246 & 1011 & 235 & 2916 & 126 & 2790 & 4.32 & 95.68 & 84.18 \\
RF (GR) & 1673 & 1042 & 631 & 2489 & 95 & 2394 & 3.82 & 96.18 & 86.03 \\
\hline
\multicolumn{10}{l}{\small \textit{Note:} Flagged = TP + FP, corresponding to footprints predicted as erroneous and therefore removed or prioritized for manual review;} \\
\multicolumn{10}{l}{\small Retained = TN + FN.}
\end{tabular}%
}
\end{table}

\section{Discussion}
The segmentation stage should be regarded as a controlled source of candidate footprint objects rather than a conclusive step in database production. Although U-Net and SAM-LoRA exhibited similar final-epoch diagnostics (\autoref{tab:tab2}), their vectorized and consolidated outputs still manifested object-level error modes as documented in Figure 5, including false positives, merged structures, fragments, incomplete outlines, and locally distorted regularized polygons. This behavior aligns with object-based remote-sensing theory: once a raster mask is transformed into a polygon layer, quality assessment must transition from pixel agreement to object-level evidence derived from footprint form, neighborhood structure, and image response (Blaschke, 2010; Ma et al., 2017; Niroshan and Carswell, 2025).
The results from the independent Site E corroborate this interpretation. Configurations that integrated geometric and contextual predictors proved more effective than those relying predominantly on contextual and raster evidence. DT (GC) achieved the highest independent-test performance, with 95.31\% accuracy, 91.06\% F1-score, 92.82\% balanced accuracy, and 0.880 MCC. These findings suggest that many segmentation errors were expressed through diagnostically interpretable object properties, such as weak compactness, low rectangular occupancy, irregular vertex structure, local area anomaly, or abnormal neighborhood density. While geometry and raster also remained competitive, spectral and texture information demonstrated less stability across varying conditions of roof color, shadow, vegetation, and local contrast. The observation that the full 24-feature setup did not surpass GC indicates that transferability was influenced more by the domains selected to capture the structural form of the error under spatially independent testing than by the sheer quantity of features (Roberts et al., 2017; Wang et al., 2023).

\subsection{Trade-off between erroneous-footprint removal and acceptable-footprint preservation}
The primary aim of the proposed workflow was not solely to enhance classification accuracy but to lessen the error load in a GIS-ready footprint database while ensuring minimal loss of valid buildings. This distinction is crucial because a model with high recall might eliminate more incorrect polygons but could also remove numerous valid footprints. It will eventually increase manual recovery efforts and diminish the database's utility. The DT (GC) model offered the most balanced compromise. It lowered the residual error rate to 4.62\%, maintained 98.31\% of valid footprints, and resulted in a flagged set where 95.11\% of the objects were genuinely erroneous. These matrices are all derived from the same independent-test confusion matrix. The comparison with RF (GR) highlights this trade-off effectively. That setup identified a greater portion of erroneous footprints and achieved the lowest residual error rate, but it also led to significantly more false positive removals than the preferred model. For operational quality control, such an approach might be beneficial when the focus is on aggressive error elimination before manual verification. However, when the objective is to preserve a dependable and minimally depleted footprint database, the preferred DT (GC) configuration is more justifiable as it combines high error detection with robust preservation of valid footprints.

\subsection{Implications for post-segmentation GeoAI workflows}
The findings indicate that geometric regularization and the consolidation of cross-model footprints under spatial-exclusivity constraints alone were inadequate for achieving operational footprint readiness. In contrast, object-level ML screening led to noticeable improvements at the database level across the configurations tested. Initially, the independent consolidated Site E layer had 27.32\% erroneous footprints and a purity of 72.68\% for acceptable footprints. The grouped feature-domain configurations reduced the proportion of retained errors to between 4.26\% and 8.69\%, while increasing the purity of retained data to between 91.31\% and 95.74\% after the screening. This corresponds to relative error reductions ranging from 68.20\% to 84.41\%. In the top-performing configurations, residual errors were kept below 5\% (3.82-4.74\%), and retained purity surpassed 95\%, demonstrating that the workflow enhanced the operational usability of the consolidated segmentation-derived footprint database, rather than merely boosting classifier-level scores.\\
The results further suggest that the workflow should be viewed as a configurable quality-control filter, not as a universally fixed removal rule. More aggressive configurations, such as RF (GR), achieved the lowest residual error but also flagged a significantly higher number of acceptable footprints. In contrast, Decision Tree with Geometry and Context maintained a more balanced approach between error elimination and footprint preservation. Therefore, conservative settings are more appropriate for refining map-ready databases, while aggressive settings might be beneficial when the flagged layer is subject to further manual review. The spatially independent test design reinforces this interpretation, as Site E was not included in model fitting, feature-set selection, class-ratio adjustment, or decision-threshold tuning.

\subsection{Limitations and future research}
When evaluating the findings, it is important to acknowledge several constraints. The workflow was limited to existing building-footprint polygons derived from segmentation, which were preserved after consolidating footprints across models under spatial-exclusivity control. As a result, complete omission errors, non-building object classes, and discarded model-specific polygons were not included in the current object-level classification scope. Reference labels and representative selections across models were determined through visual interpretation of orthophotos, which is suitable for assessing footprint quality but may involve subjective judgement in cases that are partially obscured or ambiguous. Furthermore, the feature domains were predefined and varied in the number of descriptors, and all experiments utilized UAV RGB data from a single regional context. Therefore, the results at the domain level should be viewed as evidence from the tested feature-composition design rather than as a claim of feature-count normalization or cross-sensor generality.
Future research should focus on three key extensions. Firstly, the binary acceptable/error label could be expanded into a multiclass error taxonomy to differentiate false positives, merged objects, fragments, incomplete boundaries, and geometric distortions. The same quality-control logic could also be applied to other segmentation-derived geospatial object layers. Secondly, more transferable feature sets should be developed through feature selection, grouped importance analysis, equal-size subset testing, and incorporating additional data sources such as DSM/LiDAR height, NIR or thermal imagery, shadow/vegetation indicators, and segmentation uncertainty. Thirdly, graph neural networks or graph attention networks offer a particularly promising avenue, as footprints can be represented as spatially connected objects. This allows for the direct learning of adjacency, proximity, overlap, orientation, and neighborhood-density relations, rather than relying solely on tabular descriptors.

\section{Conclusion}
This study provided a GeoAI framework for deriving precise building footprint from high resolution UAV images. Cutting edge deep learning architectures like U-Net and Transformers produced building footprints lack quality to directly insert into a vector geodatabase. Our framework utilized 24 domain-aware features to improve the geometric precision of footprints and make it ready to use. The decision tree with geometric and contextual features demonstrated strong generalization on the unseen test site, achieving an accuracy of 95.31\%, an F1-score of 91.06\%, and an MCC of 0.880. At the database level, the approach effectively identified 87.34\% of erroneous footprints while preserving 98.31\% of acceptable structures, yielding an 83.09\% relative error reduction and enhancing final database purity to 95.38\%. Overall, this study demonstrates that post-segmentation, object-level ML offers a robust, highly transferable mechanism for automated quality control in production-scale geospatial databases.

\nocite{*}
\bibliography{references}

\appendix
\section{Additional Tables and Figures}

\setcounter{table}{0}
\renewcommand{\thetable}{A.\arabic{table}}

\begin{table}[htbp]
\centering
\caption{Considered decision rules for cross-model footprint consolidation under the spatial-exclusivity constraint.}
\label{tab:A1}

\begin{tabular}{p{0.42\linewidth} p{0.50\linewidth}}
\hline
\textbf{Cross-model condition} & \textbf{Retained object} \\
\hline

One acceptable and one erroneous candidate, regardless of source model
&
Retain the acceptable footprint.
\\[4pt]

Both candidates acceptable
&
Retain the footprint with stronger boundary alignment, completeness, separation from neighbouring buildings, and geometric plausibility.
\\[4pt]

Both candidates erroneous
&
Retain the less distorted and more representative erroneous footprint.
\\[4pt]

Detected by only one model
&
Retain the single available footprint.
\\

\hline
\end{tabular}

\end{table}

\renewcommand{\thetable}{A.\arabic{table}}

\begin{table}[htbp]
\centering
\caption{Site-wise sample composition after cross-model footprint consolidation under the spatial-exclusivity constraint, merged-layer preparation, and, for Sites B--D, controlled acceptable-class sampling.}
\label{tab:A2}

\small
\begin{tabular}{p{0.10\linewidth} 
                p{0.18\linewidth} 
                p{0.18\linewidth} 
                p{0.18\linewidth} 
                p{0.25\linewidth}}
\hline
\textbf{Site/partition} &
\textbf{Post-consolidation pool $n$ (A:Er)} &
\textbf{Final analytical set $n$ (A:Er)} &
\textbf{Dataset role} &
\textbf{Sampling treatment} \\
\hline

Site B &
3,183 (2,793:390) &
1,451 (1,061:390) &
ML training &
Controlled A-class undersampling
\\[4pt]

Site C &
2,997 (2,030:967) &
1,745 (778:967) &
ML training &
Controlled A-class undersampling
\\[4pt]

Site D &
10,543 (8,187:2,356) &
5,502 (3,146:2,356) &
ML training &
Controlled A-class undersampling
\\[4pt]

Sites B--D &
16,723 (13,010:3,713) &
8,698 (4,985:3,713) &
Final ML training set &
Controlled A-class undersampling
\\[4pt]

Site E &
4,162 (3,025:1,137) &
4,162 (3,025:1,137) &
Independent testing &
No class-ratio adjustment
\\[4pt]

Total &
20,885 (16,035:4,850) &
12,860 (8,010:4,850) &
Final analytical dataset &
Site-specific treatment
\\
\hline
\end{tabular}

\vspace{2mm}

\begin{minipage}{0.97\linewidth}
\footnotesize
\textit{Note:} A = acceptable; Er = erroneous. 
All sites underwent manual labelling, cross-model footprint consolidation 
under the spatial-exclusivity constraint, and merged-layer preparation. 
Controlled acceptable-class sampling was applied only to Sites B--D.
\end{minipage}

\end{table}

\renewcommand{\thetable}{A.\arabic{table}}

\begin{table}[htbp]
\centering
\caption{Confusion-matrix outcomes used for independent footprint-quality evaluation.}
\label{tab:A3}

\small
\begin{tabular}{
p{0.20\linewidth}
p{0.34\linewidth}
p{0.38\linewidth}}
\hline
\textbf{Outcome} &
\textbf{Definition} &
\textbf{Operational interpretation} \\
\hline

True positive (TP) &
Erroneous footprint predicted as erroneous &
Error correctly flagged for removal or review
\\[4pt]

True negative (TN) &
Acceptable footprint predicted as acceptable &
Valid footprint correctly retained
\\[4pt]

False positive (FP) &
Acceptable footprint predicted as erroneous &
Valid footprint unnecessarily flagged
\\[4pt]

False negative (FN) &
Erroneous footprint predicted as acceptable &
Residual error retained in the database
\\

\hline
\end{tabular}

\end{table}

\setcounter{figure}{0}
\renewcommand{\thefigure}{A.\arabic{figure}}

\begin{figure}[htbp]
  \centering
  \includegraphics[width=0.70\linewidth]{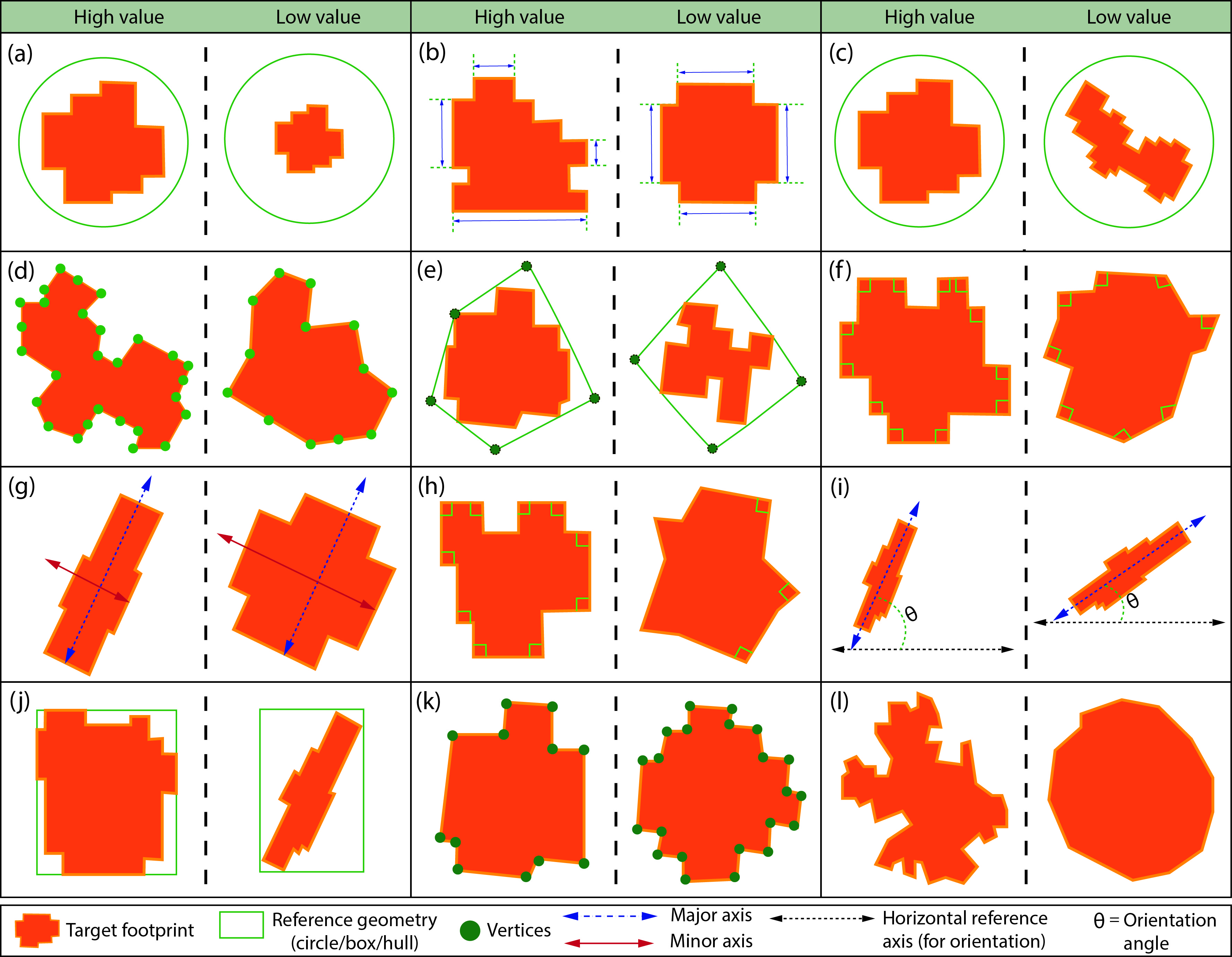}
  \caption{Conceptual high- and low-value configurations of the 12 geometric footprint descriptors.}
  \label{fig:A1}
\end{figure}

\renewcommand{\thefigure}{A.\arabic{figure}}
\begin{figure}[htbp]
  \centering
  \includegraphics[width=0.70\linewidth]{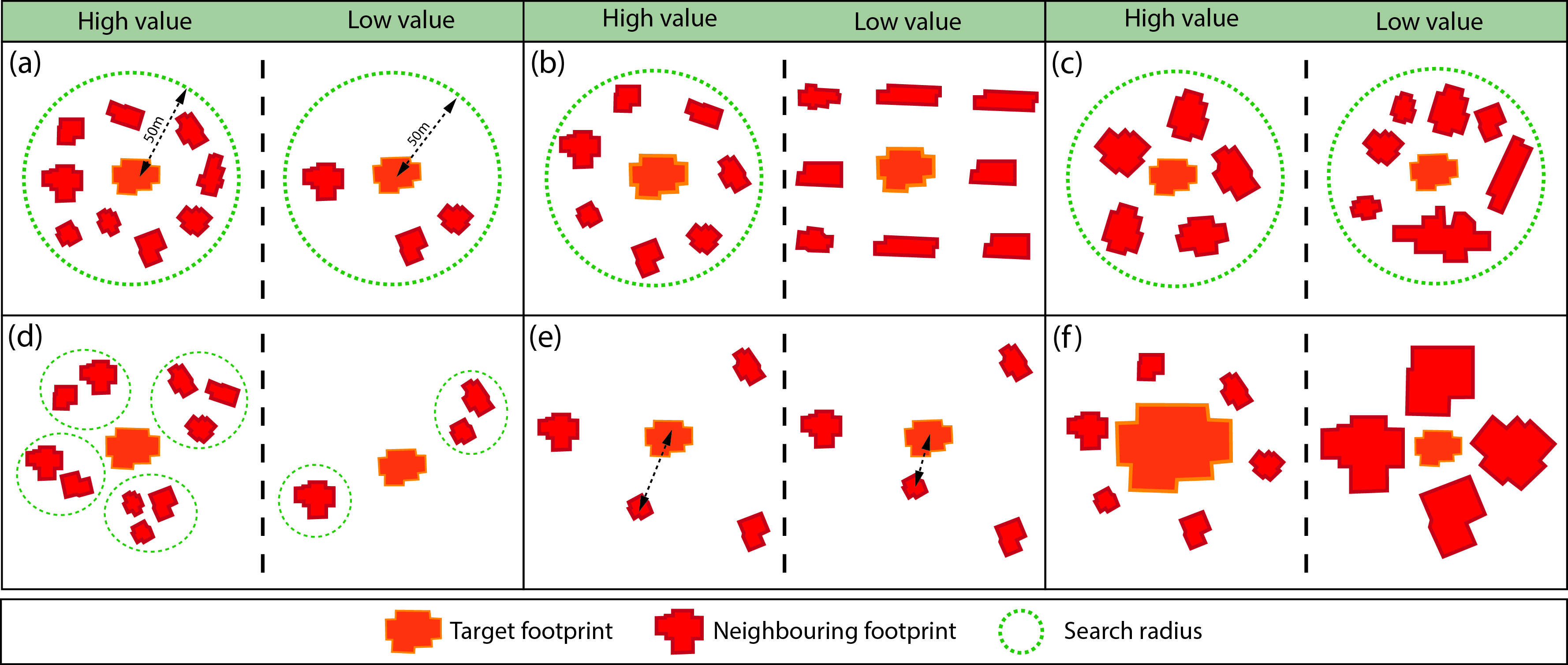}
  \caption{Conceptual high- and low-value configurations of the six spatial-contextual descriptors. The focal footprint is orange, neighboring footprints are red, and dashed circles indicate search neighborhoods.}
  \label{fig:A2}
\end{figure}

\renewcommand{\thefigure}{A.\arabic{figure}}
\begin{figure}[htbp]
  \centering
  \includegraphics[width=0.70\linewidth]{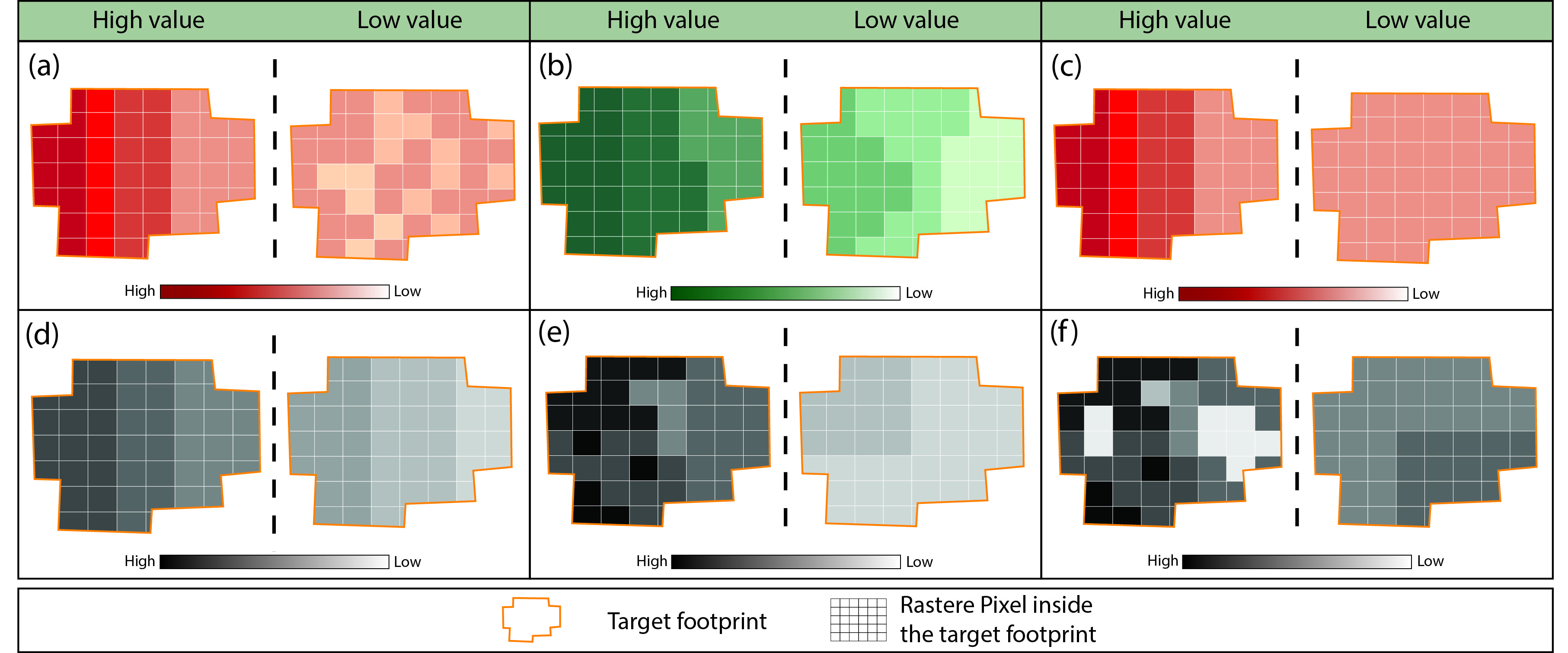}
  \caption{Conceptual high- and low-value configurations of the six raster-derived descriptors; statistics were calculated within the target footprint.}
  \label{fig:A3}
\end{figure}

\end{document}